\documentclass{article}

 \usepackage[dblblindworkshop,final]{neurips_2026}
 \workshoptitle{AXIOM Foundations of Efficient Deep Learning}

\usepackage[utf8]{inputenc} 
\usepackage[T1]{fontenc}    
\usepackage{hyperref}       
\usepackage{url}            
\usepackage{booktabs}       
\usepackage{amsfonts}       
\usepackage{nicefrac}       
\usepackage{microtype}      
\usepackage{xcolor}         
\usepackage{graphicx}       
\usepackage{amsmath}
\usepackage{amssymb}
\usepackage{multirow}
\usepackage{makecell}
\usepackage{tikz}
\usepackage{pgfplots}

\title{Probing SAM Perturbations Through Statistical Replay}
\workshoptitle{AXIOM 2026: Foundations of Efficient Deep Learning}

\author{%
  Ting Han\textsuperscript{1,\,2,\,3} \quad Jens Kleesiek\textsuperscript{1,\,3} \quad Michael Kamp\textsuperscript{1,\,2,\,3} \\
  \textsuperscript{1}TU Dortmund, Germany \\
  \textsuperscript{2}Lamarr Institute for Machine Learning and Artificial Intelligence, Germany\\
  \textsuperscript{3}Institute for Artificial Intelligence in Medicine, UK Essen, Germany\\
  \texttt{ting.han@tu-dortmund.de} \\
}

\begin{document}

\maketitle

\begin{abstract}

Sharpness Aware Minimization (SAM) updates model parameters using gradients evaluated at adversarially perturbed weights, but it remains unclear which aspects of this procedure are sufficient to reproduce its generalization behavior.
In our ResNet 18 experiments on CIFAR 10, simple curvature regularizers based on the Hessian trace and largest eigenvalue remain below SAM in validation accuracy, motivating a direct examination of its perturbations.
We introduce statistical replay, a diagnostic that summarizes perturbations from reference SAM trajectories using elementwise means and standard deviations over local windows and constructs new perturbations during independent SGD training. 
Pure replay uses only these summaries, whereas mixed replay adds isotropic Gaussian noise. Pure replay does not reproduce SAM's validation behavior, while mixed replay and pure isotropic perturbations can approach its validation performance at selected scales without reproducing its training dynamics.
An equal radius random direction within SAM performs similarly to SGD, and source, temporal, and layer interventions further suggest that replay behavior depends on perturbation construction.
Together, these preliminary results weaken several simple explanations of SAM while leaving an important part of its mechanism unexplained.

\end{abstract}

\section{Introduction}
\label{sec:introduction}

Sharpness-Aware Minimization (SAM) is designed to improve generalization by minimizing an approximate worst-case loss within a local parameter neighborhood, rather than optimizing only the empirical loss at the model’s current parameters ~\citep{foret2021sharpnessaware}. It was originally motivated by the geometric intuition that flat solutions, whose training loss remains low under small parameter perturbations, tend to generalize~\citep{hochreiter1997flat}.

At each training step, SAM first computes an approximate adversarial parameter perturbation that increases the minibatch loss within a fixed radius neighborhood, and then updates the original parameters using the gradient evaluated at the perturbed point. Although SAM often improves generalization, it remains unclear which aspects of this procedure are sufficient to reproduce its generalization behavior.

Analyses of SAM emphasize loss landscape geometry, relating its behavior to curvature regulation and movement toward flatter regions~\citep{wen2023how,bartlett2023dynamics,agarwala2023sam}. However, final solution sharpness does not consistently account for SAM's generalization behavior~\citep{vani2024forget}. Motivated by these geometric accounts, we first test simple curvature regularizers based on the Hessian trace and largest eigenvalue. In our preliminary experiments, neither reproduces SAM's validation behavior, as reported in Appendix~\ref{app:curvature}. This motivates us to examine perturbation construction more directly. Minibatch stochastic gradient noise within SAM perturbations can play an important role in generalization~\citep{Li_2024_CVPR}. Random parameter perturbations provide only marginal improvements in some settings~\citep{andriushchenko2022towards}, whereas carefully constructed random weight perturbations can achieve strong generalization performance~\citep{li2024revisiting}. The allocation of SAM perturbations across model layers can also affect optimization~\citep{haas2024boldsymbolmumathbfp}. These studies analyze or modify perturbations within the training trajectory being optimized. Our work takes a complementary diagnostic perspective by examining which information remains useful when local statistics of recorded SAM perturbations are transferred to an independent trajectory.

To study this question, we treat SAM's adversarial parameter perturbation, $\Delta_t=w_t^{\mathrm{perturbed}}-w_t$, as a \emph{diagnostic object}. We record perturbations from reference SAM trajectories and summarize them using elementwise means and standard deviations over local training windows. During independent SGD training, we sample new perturbations from these summaries and apply them during gradient evaluation, a procedure we call \emph{statistical replay}. This discards the exact recorded directions and their coupling to the source trajectory. We examine two variants: \emph{pure statistical replay} uses only the recorded summaries, whereas \emph{mixed statistical replay} supplements their elementwise variation with an isotropic Gaussian component. To determine whether the effect of mixed replay requires the recorded statistics, we also compare it with pure isotropic perturbations. We vary the source and temporal assignment of the recorded statistics and the layers receiving the perturbations. Replay examines what remains useful outside SAM, but it does not determine whether an arbitrary direction is sufficient within the original SAM update. We therefore include a complementary control that substitutes an equal radius random direction while leaving the remaining procedure unchanged. These interventions are intended as preliminary diagnostics rather than proposed replacements or a complete explanation of SAM.

In experiments using ResNet 18 on CIFAR 10, pure statistical replay maintains nearly perfect training accuracy but does not reproduce SAM's validation performance. Mixed replay produces stronger regularization, while suitably scaled pure isotropic perturbations achieve comparable validation performance without using recorded SAM statistics. Neither alternative reproduces SAM's combination of late validation performance and nearly perfect training accuracy. Source, temporal, and layer interventions further suggest that the effect of the recorded statistics depends on replay construction. Mixed replay is more consistently sensitive to source and temporal alignment, while layer restriction does not improve pure replay and some broad residual layer groups maintain mixed replay performance. The equal radius random direction control performs similarly to SGD and does not reproduce SAM's validation behavior. Taken together, our preliminary curvature regularization and perturbation experiments weaken several simple explanations of SAM in the setting studied here, while leaving an important part of its mechanism unexplained.

\section{Controlled Analysis of SAM Perturbations}
\label{sec:method}

\paragraph{SAM perturbations and random direction control.}
For minibatch $\mathcal{B}_t$, standard nonadaptive SAM computes
\begin{equation}
\Delta_t
=
\rho
\frac{
\nabla_w \mathcal{L}_{\mathcal{B}_t}(w_t)
}{
\left\lVert
\nabla_w \mathcal{L}_{\mathcal{B}_t}(w_t)
\right\rVert_2
}.
\end{equation}
For the random direction control, we replace $\Delta_t$ with
\begin{equation}
\Delta_t^{\mathrm{rand}}
=
\rho
\frac{z_t}{\lVert z_t\rVert_2},
\qquad
z_t \sim \mathcal{N}(0,I),
\end{equation}
The first gradient evaluation is retained, but the normalized gradient
direction is replaced by the random direction above. All other
components of SAM remain unchanged.

Along a reference SAM trajectory, we record $\Delta_t$ and compute its
elementwise mean $\mu_{e,j}$ and standard deviation $\sigma_{e,j}$
separately for each parameter tensor. By default, statistics are
computed over consecutive windows of $K=10$ minibatches. 

\paragraph{Statistical replay.}
During an independent SGD run, we sample
\begin{equation}
\xi_{e,j}
=
\beta\mu_{e,j}
+
\alpha
\left[
(1-\lambda_{\mathrm{mix}})
\sigma_{e,j}\odot z
+
\lambda_{\mathrm{mix}}
\sigma_{\mathrm{iso}}\eta
\right],
\end{equation}
where $z,\eta\sim\mathcal{N}(0,I)$ are sampled independently for each
minibatch. The parameters $\beta$ and $\alpha$ scale the recorded mean
and random components, respectively, while $\lambda_{\mathrm{mix}}$
controls the contributions of the recorded variation and isotropic
noise. The parameter $\sigma_{\mathrm{iso}}$ sets the scale of the
isotropic component. We refer to $\lambda_{\mathrm{mix}}=0$ as
\emph{pure statistical replay} and $\lambda_{\mathrm{mix}}>0$ as
\emph{mixed statistical replay}. We use
$\xi_t^{\mathrm{iso}}=\alpha\sigma_{\mathrm{iso}}\eta_t$ as the pure
isotropic baseline. As in SAM, the sampled perturbation is temporarily
added to the current parameters. After evaluating the gradient at the
perturbed parameters, we restore the original parameters and use the
gradient for the SGD update.

\section{Results}
\label{sec:results}

\begin{table*}[t]
  \caption{
    Perturbation experiments on CIFAR-10, averaged over three seeds.
    Late Train and Late Val.\ are averaged over epochs 201--250.
    Late $N_{80}$ is the percentage of these epochs with validation
    accuracy $\geq 80\%$. The first block compares SAM with its random
    direction control under matched weight decay. For pure and mixed replay,
    parentheses denote $(\alpha,\beta)$. Mixed replay uses
    $(\lambda_{\mathrm{mix}},\sigma_{\mathrm{iso}})=(0.10,0.40)$.
    For isotropic perturbations, parentheses denote
    $(\alpha,\sigma_{\mathrm{iso}})$.
  }
  \label{tab:main-results}
  \centering

  \small
  \setlength{\tabcolsep}{4pt}
  \renewcommand{\arraystretch}{1.05}

  \begin{tabular}{lcccc}
    \toprule
    Method
    & WD
    & Late Train
    & Late Val.
    & Late $N_{80}$ (\%) \\
    \midrule

    Standard SAM
    & $2\times10^{-4}$
    & 99.996
    & 80.474
    & 93.33 \\

    Random direction SAM
    & $2\times10^{-4}$
    & 99.031
    & 76.475
    & 0.00 \\

    \midrule

    Standard SGD
    & $2\times10^{-4}$
    & 99.032
    & 76.693
    & 0.00 \\

    Standard SGD
    & $0$
    & 99.959
    & 78.077
    & 0.00 \\

    \midrule

    Isotropic $(1.25,0.05)$
    & $0$
    & 98.483
    & 79.523
    & 6.67 \\

    Isotropic $(1.50,0.05)$
    & $0$
    & 95.665
    & 80.109
    & 64.00 \\

    Isotropic $(1.75,0.05)$
    & $0$
    & 89.845
    & 79.610
    & 12.67 \\

    \midrule

    Pure replay $(1.25,1.50)$
    & $0$
    & 99.964
    & 78.486
    & 0.00 \\

    Pure replay $(1.50,1.00)$
    & $0$
    & 99.966
    & 78.371
    & 0.00 \\

    Pure replay $(1.75,1.50)$
    & $0$
    & 99.981
    & 78.637
    & 0.00 \\

    \midrule

    Mixed replay $(1.25,1.50)$
    & $0$
    & 99.405
    & 79.771
    & 18.00 \\

    Mixed replay $(1.50,1.00)$
    & $0$
    & 98.706
    & 79.832
    & 14.67 \\

    Mixed replay $(1.75,1.50)$
    & $0$
    & 96.908
    & 80.118
    & 69.33 \\

    \bottomrule
  \end{tabular}
\end{table*}

Our main experiments use ResNet-18 on CIFAR-10. Unless otherwise
stated, replay uses statistics computed over consecutive windows of
$K=10$ minibatches. Table~\ref{tab:main-results} compares standard SAM
with random direction SAM under matched weight decay and evaluates pure
statistical replay, mixed statistical replay, and isotropic
perturbations against SGD. Table~\ref{tab:structure_main} examines
temporal ordering and layer selection for pure and mixed statistical replay.
Additional experiments on source replacement,
weight decay, epoch aggregation, and replay component ablations,
together with full experimental settings, hyperparameters, and seed
level results are in Appendix~\ref{app:experimental_details}. The tested curvature regularizers reach $79\%$ validation accuracy more
frequently than SGD but remain substantially below SAM
(Appendix~\ref{app:curvature}).

\paragraph{Matched direction control.} Under matched weight decay, replacing the SAM ascent direction with an
equal radius random direction reduces late validation accuracy by
approximately $4\%$. Random direction SAM performs
similarly to SGD, and neither method reaches $80\%$ validation accuracy, whereas standard SAM maintains this level
for most of this period. Thus, an arbitrary direction with the same
radius does not reproduce the generalization benefit of SAM.

\paragraph{Pure replay, mixed replay, and isotropic perturbations.} With weight decay $0$, pure statistical replay gives only a modest
increase in late validation accuracy over SGD while maintaining nearly
perfect training accuracy. None of the pure replay configurations
reaches $80\%$ validation accuracy at the late training. Mixed
replay produces stronger regularization, with some configurations
maintaining validation accuracy above $80\%$ with reduced training performance. A similar pattern is observed for isotropic perturbations.
The recorded means and standard deviations alone do not reproduce
the stronger regularization of mixed replay, and neither mixed replay
nor suitably scaled isotropic perturbations consistently outperforms
the other.

\paragraph{Temporal ordering.} Table~\ref{tab:structure_main}(a) examines two forms of temporal
alignment for pure and mixed statistical replay. For pure replay,
permuting the batch windows within each source epoch reduces late
validation accuracy across all three configurations, whereas permuting
the source epochs has no consistent effect. For mixed replay, both forms of permutation generally reduce late
validation accuracy relative to aligned replay. A similar pattern is
observed under source replacement, as reported in
Appendix~\ref{app:source_method}.
Mixed replay is therefore more consistently sensitive to the source and
temporal alignment of the recorded statistics than pure replay.

\paragraph{Layer selection.}
Table~\ref{tab:structure_main}(b) examines layer selection for pure and
mixed statistical replay (full results in Appendix~\ref{app:layerwise_method}). For pure replay, perturbing all parameters
gives the highest late validation accuracy, while restricting the
perturbations to the tested layer groups slightly reduces validation
accuracy without materially changing training accuracy. For mixed
replay, perturbing L1--L4 gives validation performance comparable to
perturbing all parameters. In contrast, restricting mixed replay to the
classifier substantially reduces late validation accuracy and restores nearly perfect training accuracy. Thus, the
effect of layer selection depends on the construction of the replay
perturbation.

\begin{table*}[t]
  \caption{
    Temporal ordering and layer selection for pure and mixed statistical
    replay with weight decay $0$, averaged over three seeds. Train and
    Val.\ are averaged over epochs 201--250. Mixed replay uses
    $(\lambda_{\mathrm{mix}},\sigma_{\mathrm{iso}})=(0.10,0.40)$ throughout.
    Aligned results for the temporal comparison are reported in
    Table~\ref{tab:main-results}. In \textbf{(a)}, Batch permutes the batch
    windows within each epoch, while Epoch permutes the source epochs.
    Layer selection in \textbf{(b)} uses
    $(\alpha,\beta)=(1.75,1.50)$.
  }
  \label{tab:structure_main}
  \centering

   \begin{minipage}[t]{0.56\textwidth}
    \centering
    \textbf{(a) Temporal ordering}

    \vspace{0.5em}

    \small
    \setlength{\tabcolsep}{2pt}
    \renewcommand{\arraystretch}{1.05}
    \begin{tabular}{llcccc}
      \toprule
      &
      &
      \multicolumn{2}{c}{Pure replay}
      & \multicolumn{2}{c}{Mixed replay} \\
      \cmidrule(lr){3-4}
      \cmidrule(lr){5-6}
      Ordering
      & $(\alpha,\beta)$
      & Train
      & Val.
      & Train
      & Val. \\
      \midrule

      \multirow{3}{*}{Batch}
      & $(1.25,1.50)$
      & 99.981 & 78.066
      & 99.391 & 79.622 \\

      & $(1.50,1.00)$
      & 99.976 & 78.278
      & 98.726 & 79.529 \\

      & $(1.75,1.50)$
      & 99.965 & 78.281
      & 97.230 & 79.281 \\

      \midrule

      \multirow{3}{*}{Epoch}
      & $(1.25,1.50)$
      & 99.972 & 77.897
      & 99.388 & 79.601 \\

      & $(1.50,1.00)$
      & 99.985 & 78.674
      & 98.700 & 79.609 \\

      & $(1.75,1.50)$
      & 99.969 & 78.626
      & 97.233 & 79.588 \\

      \bottomrule
    \end{tabular}
  \end{minipage}%
  \hfill
  \begin{minipage}[t]{0.43\textwidth}
    \centering
    \textbf{(b) Layer selection}

    \vspace{0.5em}

    \small
    \setlength{\tabcolsep}{2pt}
    \renewcommand{\arraystretch}{1.05}
    \begin{tabular}{lcccc}
      \toprule
      &
      \multicolumn{2}{c}{Pure replay}
      & \multicolumn{2}{c}{Mixed replay} \\
      \cmidrule(lr){2-3}
      \cmidrule(lr){4-5}
      Layers
      & Train
      & Val.
      & Train
      & Val. \\
      \midrule

      All
      & 99.981 & 78.637
      & 96.908 & 80.118 \\

      L1-L4
      & 99.961 & 78.415
      & 97.983 & 80.177 \\

      L1-L4 $+$ FC
      & 99.978 & 78.246
      & 97.782 & 79.734 \\

      FC only
      & 99.973 & 78.117
      & 99.957 & 76.002 \\

      \bottomrule
    \end{tabular}
  \end{minipage}
\end{table*}


\section{Discussion and Conclusion}
\label{sec:discussion}

In the setting studied here, the tested trace and largest eigenvalue
regularizers generally improve validation over SGD but remain
substantially below SAM (Appendix~\ref{app:curvature}). The replay experiments provide a complementary result. Pure statistical replay produces only modest regularization, whereas mixed replay produces substantially stronger effects. At selected scales, pure isotropic perturbations achieve validation performance comparable to mixed replay without using recorded SAM statistics, suggesting that these statistics are not uniquely required for such improvements. When their validation performance approaches SAM, mixed replay and isotropic perturbations exhibit more slowly increasing and lower training accuracy, with weight decay further widening this gap (Appendix~\ref{app:weight_decay_results}). The equal radius random direction control also performs similarly to SGD rather than SAM, suggesting that an arbitrary direction is insufficient within the original update.

These experiments do not identify a complete mechanism for SAM. Our experimental scope is limited to ResNet 18 on CIFAR 10 without data augmentation. The curvature controls are limited to trace and largest eigenvalue penalties on the classifier head, while replay effectiveness depends on the scale and composition of its random component. The strongest mixed replay and isotropic results therefore show that validation performance comparable to SAM is achievable at selected settings, not that it is robust across hyperparameter choices. The effects of replay composition, temporal organization, and layer support may also differ across architectures, datasets, augmentation settings, and SAM variants.

Overall, these preliminary results weaken explanations based solely on the tested curvature penalties, perturbation radius, elementwise perturbation statistics, or tested restrictions of layer support. They leave unresolved which directional correlations, minibatch dependencies, trajectory effects, or interactions between SAM’s two gradient evaluations account for its generalization behavior.

\newpage

\bibliography{references}
\bibliographystyle{plainnat}

\newpage
\appendix
\onecolumn

\section{Acknowledgement}
\label{sec:ack}
This research work is supported by the Federal Ministry of Education and Research of Germany and the state of North Rhine-Westphalia as part of the Lamarr Institute for Machine Learning and Artificial Intelligence, and by the German Federal Ministry of Research, Technology and Space (DECIPHER-M, 01KD2420C).

\section{Related Work}
\label{sec:related_work}

\paragraph{SAM and generalization.}
Sharpness-Aware Minimization (SAM) optimizes neighborhood loss using
gradients evaluated at adversarially perturbed parameters
\citep{foret2021sharpnessaware}. Its behavior has been studied through
curvature and optimization dynamics: SAM has been connected to
alignment between gradients and leading Hessian eigendirections,
dynamical regulation of large Hessian eigenvalues, and trajectory
dependent motion toward lower curvature regions
\citep{wen2023how,agarwala2023sam,bartlett2023dynamics}.
However, final solution sharpness does not consistently predict
generalization across practical settings, and sharpness minimizing
algorithms can generalize through effects beyond sharpness reduction
\citep{pmlr-v202-andriushchenko23a,wen2023sharpness}. Other analyses explain
SAM through its implicit bias or the perturbation induced forgetting
of undesirable model biases
\citep{andriushchenko2022towards,vani2024forget}.

\paragraph{Structure and stochasticity of parameter perturbations.}
Prior work studies how perturbation direction, randomness, temporal
reuse, and parameter support affect SAM and related training
procedures. Random parameter directions provide limited improvements
in some settings \citep{andriushchenko2022towards}, whereas
Random Sharpness-Aware Minimization
instead uses Gaussian initialization before adversarial ascent
\citep{liu2022random}, and random weight perturbation shows that
suitably scaled stochastic parameter noise can itself provide strong
regularization \citep{li2024revisiting}. These methods differ in
whether randomness replaces or supplements the adversarial direction
and whether the SAM ascent step is retained. Friendly SAM examines the
contribution of minibatch specific stochastic gradient noise
\citep{Li_2024_CVPR}. LookSAM periodically computes the SAM update
and reuses its component orthogonal to the current SGD gradient across
nearby optimization steps \citep{liu2022towards}. Within the original
SAM update, sparse and normalization only variants restrict which
parameters are perturbed, while layerwise analyses modify perturbation
scaling
\citep{mi2022make,mueller2023normalization,haas2024boldsymbolmumathbfp}.
Our work takes a complementary diagnostic perspective: we summarize
perturbations recorded from reference SAM trajectories and replay them
during independent SGD training while varying their source, temporal
assignment, and layer support.

\section{Method and Implementation Details}
\label{app:experimental_details}

\paragraph{Dataset and model.}
All main experiments use CIFAR-10 with ResNet-18 trained from scratch.
We use the standard ImageNet-style ResNet-18 architecture, retaining the
$7\times7$ first convolutional layer and the initial max-pooling layer.
No data augmentation is used.

\paragraph{Optimization.}
All trajectories are trained for 250 epochs with batch size 64.
All methods use a learning rate of $0.01$, momentum $0.9$, and no
learning-rate scheduler.
Standard SGD and replay trajectories use weight decay $0$.
Reference SAM trajectories use perturbation radius $\rho=0.09$ and
weight decay $2\times10^{-4}$. All experiments were run on NVIDIA H100 GPUs, each with 80\,GB of memory.

\paragraph{Random seeds.}
Unless otherwise stated, multi-seed results use Seeds $1$, $42$, and
$15213$. Single-seed intervention experiments use Seed $42$ unless stated
otherwise.

\paragraph{Default perturbed layers.}
Unless otherwise stated, replay perturbations are applied to all trainable
parameters. Layer-wise experiments explicitly restrict the set of parameters
receiving the perturbation.

\paragraph{Representative replay configurations.}
We first evaluate 375 mixed replay configurations using Seed 42, with $\alpha,\beta\in{1.00,1.25,1.50,1.75,2.00}$, $\lambda_{\mathrm{mix}}\in{0.10,0.20,0.30}$, and $\sigma_{\mathrm{iso}}\in{0.10,0.20,0.30,0.40,0.50}$. From this sweep, we select three representative configurations that achieve relatively high late validation accuracy while spanning different levels of late training accuracy: $(1.25,1.50,0.10,0.40)$, $(1.50,1.00,0.10,0.40)$, and $(1.75,1.50,0.10,0.40)$, where each tuple denotes $(\alpha,\beta,\lambda_{\mathrm{mix}},\sigma_{\mathrm{iso}})$. We evaluate these configurations using Seeds 1, 42, and 15213. The corresponding pure replay experiments use the same $(\alpha,\beta)$ values with $\lambda_{\mathrm{mix}}=0$. Additional configurations are specified in the corresponding experiments.

\paragraph{Batch-window statistics.}
The main experiments use SAM perturbations recorded at every minibatch of a
reference SAM trajectory. Each source epoch is divided into consecutive,
non-overlapping windows of $K=10$ minibatches. If the number of minibatches
is not divisible by $K$, the final window contains the remaining minibatches.
For each parameter tensor, we compute the elementwise mean and standard
deviation of the recorded perturbations within each window, denoted by
$\mu_{e,j}$ and $\sigma_{e,j}$ for source epoch $e$ and window $j$. 

In aligned batch-window replay, target epoch $e$ and target window $j$ use
the corresponding statistics $(\mu_{e,j},\sigma_{e,j})$. The same statistics
are used for all target minibatches within that window, while the Gaussian
variables in the replay perturbation are resampled independently for every
minibatch. Replay uses only these stored statistics and does not reuse the
individual recorded SAM perturbations. Results for individual seeds for the mixed and pure replay configurations reported in Table~\ref{tab:main-results} are provided in Table~\ref{tab:replay-seeds} and Table~\ref{tab:pure-replay-seeds}, respectively.

\paragraph{Replay implementation.}
\label{app:statistical_replay}
The replay perturbation and update rule are defined in
Section~\ref{sec:method}. For each target minibatch, the sampled replay
perturbation is temporarily added to the current parameters. The gradient is
evaluated at the perturbed parameters, after which the perturbation is removed
and the gradient is used to update the original parameters.

Replay does not perform the SAM ascent step on the target trajectory.
Therefore, even aligned replay does not reproduce the original SAM
perturbation sequence or the corresponding SAM training trajectory. 

For source replacement, all replay runs use Seed 42 for SGD training, including model initialization, minibatch ordering, and perturbation sampling. Aligned replay uses statistics recorded from the Seed 42 SAM trajectory, whereas cross source replay uses statistics from the Seed 1 or Seed 15213 SAM trajectory. Thus, each matched comparison changes only the SAM trajectory supplying the recorded means and standard deviations.

For batch window permutation, the recorded windows within each source epoch are randomly reassigned to the SGD windows of the corresponding epoch. Each source window is used exactly once, while the source epoch and the recorded statistics themselves remain unchanged. The permutation for epoch \(e\) is generated by a separate local random number generator with seed \(42+e\), where \(e\) is zero indexed. This local generator does not alter the global random states controlling SGD training and perturbation sampling.

For epoch permutation, we generate one permutation of the recorded SAM epochs using a separate local random number generator with Seed 42. During SGD epoch \(e\), replay uses all windows from the assigned SAM epoch in their original order. Each SAM epoch is used exactly once, so the intervention changes the correspondence between SAM and SGD epochs while preserving the ordering of windows within each source epoch. Generating the permutation does not alter the global random states controlling SGD training and perturbation sampling.

For each minibatch, we first perform a forward pass without gradient computation at the unperturbed parameters to update the batch normalization running statistics. We then save the updated batch normalization state, apply the replay perturbation, and compute the loss and gradients at the perturbed parameters. Before the SGD update, we restore the original parameters and the saved batch normalization state. Thus, the update uses gradients evaluated at the perturbed parameters, while the batch normalization running statistics are updated only at the unperturbed parameters.

\begin{table}[t]
    \caption{
    Seed-level results for the three representative replay configurations
    reported in Table~\ref{tab:main-results}. Train Acc.\ and Val.\ Acc.\
    are averaged over epochs 201--250. Best Val.\ is the maximum validation
    accuracy over all 250 epochs. Late $N_{80}$ is the number of epochs among
    epochs 201--250 with validation accuracy $\geq80\%$.
    }
    \label{tab:replay-seeds}
    \centering
    \begin{tabular}{lccccc}
        \toprule
        $(\alpha,\beta,\lambda_{\mathrm{mix}},\sigma_{\mathrm{iso}})$
        & Seed
        & Train Acc.
        & Val. Acc.
        & Best Val.
        & Late $N_{80}$ \\
        \midrule

        \multirow{3}{*}{$(1.25,1.50,0.10,0.40)$}
        & 1
        & 99.408 & 79.671 & 80.020 & 2/50 \\
        & 42
        & 99.397 & 79.979 & 80.290 & 25/50 \\
        & 15213
        & 99.410 & 79.665 & 80.010 & 0/50 \\

        \midrule

        \multirow{3}{*}{$(1.50,1.00,0.10,0.40)$}
        & 1
        & 98.695 & 79.839 & 80.320 & 8/50 \\
        & 42
        & 98.761 & 79.782 & 80.560 & 2/50 \\
        & 15213
        & 98.662 & 79.876 & 80.340 & 12/50 \\

        \midrule

        \multirow{3}{*}{$(1.75,1.50,0.10,0.40)$}
        & 1
        & 96.981 & 79.858 & 80.330 & 9/50 \\
        & 42
        & 96.799 & 80.272 & 80.650 & 47/50 \\
        & 15213
        & 96.944 & 80.225 & 80.560 & 48/50 \\

        \bottomrule
    \end{tabular}
\end{table}

\begin{table}[t]
    \caption{
    Seed-level results for the three representative pure replay
    configurations. Train Acc.\ and Val.\ Acc.\ are averaged over
    epochs 201--250. Best Val.\ is the maximum validation accuracy over
    all 250 epochs. Late $N_{80}$ is the number of epochs among epochs
    201--250 with validation accuracy $\geq80\%$.
    }
    \label{tab:pure-replay-seeds}
    \centering
    \begin{tabular}{lccccc}
        \toprule
        $(\alpha,\beta)$
        & Seed
        & Train Acc.
        & Val. Acc.
        & Best Val.
        & Late $N_{80}$ \\
        \midrule

        \multirow{3}{*}{$(1.25,1.50)$}
        & 1
        & 99.965 & 78.576 & 79.290 & 0/50 \\
        & 42
        & 99.965 & 78.304 & 79.010 & 0/50 \\
        & 15213
        & 99.963 & 78.579 & 79.110 & 0/50 \\

        \midrule

        \multirow{3}{*}{$(1.50,1.00)$}
        & 1
        & 99.966 & 78.565 & 79.450 & 0/50 \\
        & 42
        & 99.977 & 78.396 & 79.070 & 0/50 \\
        & 15213
        & 99.956 & 78.152 & 78.960 & 0/50 \\

        \midrule

        \multirow{3}{*}{$(1.75,1.50)$}
        & 1
        & 99.980 & 78.552 & 79.430 & 0/50 \\
        & 42
        & 99.989 & 79.036 & 79.700 & 0/50 \\
        & 15213
        & 99.973 & 78.324 & 79.300 & 0/50 \\

        \bottomrule
    \end{tabular}
\end{table}

\paragraph{Standard SAM results.}
Table~\ref{tab:sam-seeds} reports the seed-level results for standard SAM.
Across Seeds $1$, $42$, and $15213$, SAM maintains near-perfect late-stage
training accuracy, ranging from $99.992\%$ to $99.999\%$. Late-stage
validation accuracy ranges from $80.280\%$ to $80.693\%$, and Late
$N_{80}$ ranges from $45/50$ to $50/50$. These results show that the strong
late-stage validation behavior of SAM is consistent across the three seeds.
\begin{table}[t]
    \caption{
    Seed-level results for standard SAM reported in
    Table~\ref{tab:main-results}. Train Acc.\ and Val.\ Acc.\ are averaged
    over epochs 201--250. Best Val.\ is the maximum validation accuracy
    over all 250 epochs. Late $N_{80}$ is the number of epochs among
    epochs 201--250 with validation accuracy $\geq 80\%$.
    }
    \label{tab:sam-seeds}
    \centering
    \begin{tabular}{lccccc}
        \toprule
        Method
        & Seed
        & Train Acc.
        & Val. Acc.
        & Best Val.
        & Late $N_{80}$ \\
        \midrule

        \multirow{3}{*}{Standard SAM}
        & 1
        & 99.999 & 80.280 & 80.810 & 45/50 \\
        & 42
        & 99.998 & 80.693 & 81.270 & 50/50 \\
        & 15213
        & 99.992 & 80.449 & 80.920 & 45/50 \\

        \bottomrule
    \end{tabular}
\end{table}

\paragraph{Random SAM results.}
Table~\ref{tab:random-sam-seeds} reports the seed-level results for the
random-direction SAM control. Across Seeds $1$, $42$, and $15213$,
late-stage training accuracy remains close to $99\%$, while late-stage
validation accuracy stays below $77\%$. None of the three runs reaches
$80\%$ validation accuracy in the final 50 epochs. Compared with standard
SAM, replacing the ascent direction with a random direction therefore
substantially reduces validation performance while leaving training fit high.

\begin{table}[t]
    \caption{
    Seed-level results for random-direction SAM with weight decay
    $2\times10^{-4}$. Train Acc.\ and Val.\ Acc.\ are averaged over
    epochs 201--250. Best Val.\ is the maximum validation accuracy over
    all 250 epochs. Late $N_{80}$ is the number of epochs among epochs
    201--250 with validation accuracy $\geq 80\%$.
    }
    \label{tab:random-sam-seeds}
    \centering
    \begin{tabular}{lccccc}
        \toprule
        Method
        & Seed
        & Train Acc.
        & Val. Acc.
        & Best Val.
        & Late $N_{80}$ \\
        \midrule

        \multirow{3}{*}{Random-direction SAM}
        & 1
        & 99.035 & 76.377 & 78.070 & 0/50 \\
        & 42
        & 99.023 & 76.424 & 77.870 & 0/50 \\
        & 15213
        & 99.036 & 76.625 & 78.190 & 0/50 \\

        \bottomrule
    \end{tabular}
\end{table}

\paragraph{SGD results.}
Table~\ref{tab:sgd-seeds} reports the seed-level results for standard SGD
with weight decay $0$ and $2\times10^{-4}$. Without weight decay, late-stage
training accuracy remains near $100\%$, while late-stage validation accuracy
stays below $79\%$. Adding weight decay $2\times10^{-4}$ reduces both
training and validation accuracy, and none of the runs reaches $80\%$
validation accuracy in the final 50 epochs. Thus, weight decay alone does not
produce the validation behavior observed for SAM, replay, or isotropic
perturbations.
\begin{table}[t]
    \caption{
    Seed-level results for standard SGD with weight decay $0$ and
    $2\times10^{-4}$. Train Acc.\ and Val.\ Acc.\ are averaged over
    epochs 201--250. Best Val.\ is the maximum validation accuracy over
    all 250 epochs. Late $N_{80}$ is the number of epochs among epochs
    201--250 with validation accuracy $\geq 80\%$.
    }
    \label{tab:sgd-seeds}
    \centering
    \begin{tabular}{lccccc}
        \toprule
        Method
        & Seed
        & Train Acc.
        & Val. Acc.
        & Best Val.
        & Late $N_{80}$ \\
        \midrule

        \multirow{3}{*}{SGD}
        & 1
        & 99.962 & 77.878 & 78.930 & 0/50 \\
        & 42
        & 99.968 & 78.411 & 79.210 & 0/50 \\
        & 15213
        & 99.946 & 77.943 & 78.640 & 0/50 \\

        \midrule

        \multirow{3}{*}{SGD + WD}
        & 1
        & 99.036 & 76.983 & 78.150 & 0/50 \\
        & 42
        & 99.021 & 76.529 & 77.560 & 0/50 \\
        & 15213
        & 99.040 & 76.567 & 77.760 & 0/50 \\

        \bottomrule
    \end{tabular}
\end{table}

\subsection{Pure Isotropic Control}
\label{app:isotropic}

The isotropic noise is sampled as below:
\begin{equation}
\xi_{\mathrm{iso}}^{(p)}
=
\alpha
\sigma_{\mathrm{iso}}
\eta^{(p)},
\qquad
\eta^{(p)}\sim\mathcal{N}(0,I).
\end{equation}
A fresh perturbation is sampled independently for every minibatch and
applied to all trainable parameters. We do not use the SAM-derived mean or
standard deviation, and we do not apply perturbation-norm matching.

We first evaluate 60 pure isotropic configurations using Seed 42, with
\(\alpha \in \{0.75,1.00,1.25,1.50,1.75,2.00\}\) and
\(\sigma_{\mathrm{iso}} \in \{0.01,0.02,0.03,0.04,0.05,0.1,0.2,0.3,0.4,0.5\}\).
From this sweep, we select three representative configurations that
achieve relatively high late validation accuracy while spanning
different levels of late training accuracy:
\((\alpha,\sigma_{\mathrm{iso}})=(1.25,0.05)\),
\((1.50,0.05)\), and \((1.75,0.05)\).
We evaluate these configurations using Seeds 1, 42, and 15213. The seed-level results are reported in Table~\ref{tab:pure-iso-seeds}.

\begin{table}[t]
    \caption{
    Seed-level results for the three pure isotropic settings reported in
    Table~\ref{tab:main-results}. Train Acc.\ and Val.\ Acc.\ are averaged
    over epochs 201--250. Best Val.\ is the maximum validation accuracy
    over all 250 epochs. Late $N_{80}$ is the number of epochs among
    epochs 201--250 with validation accuracy $\geq80\%$.
    }
    \label{tab:pure-iso-seeds}
    \centering
    \begin{tabular}{lccccc}
        \toprule
        $(\alpha,\sigma_{\mathrm{iso}})$
        & Seed
        & Train Acc.
        & Val. Acc.
        & Best Val.
        & Late $N_{80}$ \\
        \midrule

        \multirow{3}{*}{$(1.25,0.05)$}
        & 1
        & 98.457 & 79.301 & 79.810 & 0/50 \\
        & 42
        & 98.523 & 79.859 & 80.340 & 10/50 \\
        & 15213
        & 98.470 & 79.408 & 79.980 & 0/50 \\

        \midrule

        \multirow{3}{*}{$(1.50,0.05)$}
        & 1
        & 95.930 & 80.082 & 80.430 & 36/50 \\
        & 42
        & 95.605 & 79.834 & 80.440 & 10/50 \\
        & 15213
        & 95.460 & 80.412 & 80.850 & 50/50 \\

        \midrule

        \multirow{3}{*}{$(1.75,0.05)$}
        & 1
        & 90.550 & 79.404 & 79.870 & 0/50 \\
        & 42
        & 89.857 & 79.931 & 80.380 & 19/50 \\
        & 15213
        & 89.128 & 79.496 & 79.870 & 0/50 \\

        \bottomrule
    \end{tabular}
\end{table}

Among the three configurations, $(1.50,0.05)$ gives the highest multi-seed
late-stage validation accuracy at $80.109\%$, with 96 of the 150 final
seed-epochs reaching validation accuracy at least $80\%$. Increasing the
isotropic perturbation strength reduces late-stage training accuracy:
the three-seed average decreases from $98.483\%$ for $(1.25,0.05)$ to
$95.665\%$ for $(1.50,0.05)$ and $89.845\%$ for $(1.75,0.05)$.
The corresponding late-stage validation accuracies are $79.523\%$,
$80.109\%$, and $79.610\%$, respectively. Thus, stronger isotropic
perturbations reduce training fit, while the validation improvement is
not monotonic.

Compared with standard SGD, the tested isotropic perturbations improve
late-stage validation performance and can reach the $80\%$ validation
regime. However, this improvement is associated with lower training
accuracy, especially for the stronger perturbation settings.

\subsection{Source Intervention}
\label{app:source_method}

In aligned replay, the SAM statistics are extracted from a reference SAM
trajectory with the same random seed as the SGD training run. For
cross-source replay, we keep the SGD training seed, optimization settings,
and replay hyperparameters unchanged, but use SAM statistics from a
trajectory generated with a different random seed. We evaluate the three
representative configurations for both pure and mixed replay, using Seed $42$
for SGD training and Seeds $1$ and $15213$ as alternative SAM sources. This
gives 12 cross-source trajectories in total. For mixed replay,
$(\lambda_{\mathrm{mix}},\sigma_{\mathrm{iso}})=(0.10,0.40)$. The results are presented in Table~\ref{tab:cross-source}.

For pure replay, source replacement has a configuration-dependent effect.
Relative to the aligned Seed $42$ results, late validation accuracy decreases
in three of the six cross-source comparisons and increases in the other
three. The largest reductions occur for
$(\alpha,\beta)=(1.75,1.50)$, where late validation accuracy decreases from
$79.036\%$ to $78.147\%$ and $78.107\%$. Late training accuracy remains near
$100\%$, and none of the cross-source runs reaches $80\%$ validation
accuracy during the final 50 epochs.

For mixed replay, changing the SAM source has little effect on late-stage
training accuracy but reduces validation performance relative to the aligned
Seed-$42$ results for all three representative configurations. For
$(1.25,1.50)$, aligned replay reaches $79.979\%$ late-stage validation
accuracy and Late $N_{80}=25/50$, whereas Source Seeds $1$ and $15213$
reach $79.429\%$ and $79.536\%$, respectively, with Late $N_{80}=0/50$
for both. For $(1.50,1.00)$, the aligned validation accuracy of $79.782\%$
decreases to $79.506\%$ and $79.270\%$, with neither cross-source run
reaching $80\%$ validation accuracy in the final 50 epochs. For
$(1.75,1.50)$, aligned replay reaches $80.272\%$ validation accuracy and
Late $N_{80}=47/50$, while the two cross-source runs reach $79.781\%$ and
$79.867\%$, with Late $N_{80}=4/50$ and $9/50$, respectively.

Thus, for the three representative mixed replay configurations, replacing
the aligned SAM source reduces late-stage validation performance and
high-validation persistence while leaving late-stage training accuracy
largely unchanged.
\begin{table*}[t]
    \caption{
    Cross-source replay using target Seed 42. The SAM source is either
    Seed 1 or Seed 15213. Train Acc.\ and Val.\ Acc.\ are averaged over
    epochs 201--250. Best Val.\ is the maximum validation accuracy over
    all 250 epochs. Late $N_{80}$ is the number of epochs among epochs
    201--250 with validation accuracy $\geq80\%$. For mixed replay,
    $(\lambda_{\mathrm{mix}},\sigma_{\mathrm{iso}})=(0.10,0.40)$.
    }
    \label{tab:cross-source}
    \centering
    \begin{tabular}{lllcccc}
        \toprule
        Replay
        & $(\alpha,\beta)$
        & SAM Source
        & Late Train
        & Late Val.
        & Best Val.
        & Late $N_{80}$ \\
        \midrule

        \multirow{6}{*}{Pure}
        & \multirow{2}{*}{$(1.25,1.50)$}
        & Seed 1
        & 99.987
        & 78.193
        & 78.910
        & 0/50 \\
        &
        & Seed 15213
        & 99.994
        & 78.347
        & 78.680
        & 0/50 \\
        \cmidrule(lr){2-7}

        & \multirow{2}{*}{$(1.50,1.00)$}
        & Seed 1
        & 99.984
        & 78.477
        & 79.110
        & 0/50 \\
        &
        & Seed 15213
        & 99.983
        & 78.436
        & 79.000
        & 0/50 \\
        \cmidrule(lr){2-7}

        & \multirow{2}{*}{$(1.75,1.50)$}
        & Seed 1
        & 99.960
        & 78.147
        & 78.790
        & 0/50 \\
        &
        & Seed 15213
        & 99.956
        & 78.107
        & 79.070
        & 0/50 \\

        \midrule

        \multirow{6}{*}{Mixed}
        & \multirow{2}{*}{$(1.25,1.50)$}
        & Seed 1
        & 99.385
        & 79.429
        & 79.760
        & 0/50 \\
        &
        & Seed 15213
        & 99.368
        & 79.536
        & 79.970
        & 0/50 \\
        \cmidrule(lr){2-7}

        & \multirow{2}{*}{$(1.50,1.00)$}
        & Seed 1
        & 98.695
        & 79.506
        & 79.890
        & 0/50 \\
        &
        & Seed 15213
        & 98.618
        & 79.270
        & 79.950
        & 0/50 \\
        \cmidrule(lr){2-7}

        & \multirow{2}{*}{$(1.75,1.50)$}
        & Seed 1
        & 96.971
        & 79.781
        & 80.030
        & 4/50 \\
        &
        & Seed 15213
        & 97.022
        & 79.867
        & 80.170
        & 9/50 \\

        \bottomrule
    \end{tabular}
\end{table*}


\subsection{Temporal Interventions}
\label{app:temporal_method}

We evaluate the three representative configurations for both pure and mixed
replay using Seeds $1$, $42$, and $15213$. For mixed replay,
$(\lambda_{\mathrm{mix}},\sigma_{\mathrm{iso}})=(0.10,0.40)$.
Within each source SAM epoch, the recorded batch-window statistics are
randomly permuted before being assigned to the corresponding windows of the
SGD training trajectory. The source SAM epoch and the corresponding SGD
training epoch remain aligned, and all batch-window statistics from the
source epoch are retained. Batch shuffling therefore preserves epoch-level
alignment between the SAM source trajectory and the SGD training trajectory,
while removing the original within-epoch alignment between their batch
windows. The results are reported in Table~\ref{tab:batch-shuffle-multiseed}.

Across the three pure replay configurations, batch shuffling decreases the
mean late-stage validation accuracy from $78.486\%$ to $78.066\%$,
from $78.371\%$ to $78.278\%$, and from $78.637\%$ to $78.281\%$,
respectively. Late-stage training accuracy changes little, and none of the
pure replay configurations reaches $80\%$ validation accuracy in the final
50 epochs either before or after shuffling.

For mixed replay, batch shuffling also decreases the mean late-stage
validation accuracy for all three representative configurations. The
changes are from $79.771\%$ to $79.622\%$, from $79.832\%$ to
$79.529\%$, and from $80.118\%$ to $79.281\%$, respectively. The largest
reduction occurs for $(1.75,1.50)$, for which Late $N_{80}$ decreases from
$69.33\%$ to $0.67\%$. Overall, the three-seed results show that disrupting
within-epoch window alignment reduces validation performance for both pure
and mixed replay, with a larger effect for some mixed replay configurations.

\begin{table*}[t]
    \caption{
    Comparison of aligned and batch-shuffled replay over Seeds $1$, $42$,
    and $15213$. Results are reported as mean $\pm$ sample standard deviation.
    Train Acc.\ and Val.\ Acc.\ are averaged over epochs 201--250.
    Best Val.\ is the maximum validation accuracy over all 250 epochs.
    Late $N_{80}$ is the percentage of epochs among epochs 201--250 with
    validation accuracy $\geq80\%$. For mixed replay,
    $(\lambda_{\mathrm{mix}},\sigma_{\mathrm{iso}})=(0.10,0.40)$.
    }
    \label{tab:batch-shuffle-multiseed}
    \centering
    \small
    \setlength{\tabcolsep}{3.5pt}
    \begin{tabular}{lllcccc}
        \toprule
        Replay
        & $(\alpha,\beta)$
        & Setting
        & Train Acc.
        & Val. Acc.
        & Best Val.
        & Late $N_{80}$ (\%) \\
        \midrule

        \multirow{6}{*}{Pure}
        & \multirow{2}{*}{$(1.25,1.50)$}
        & Aligned
        & $99.964 \pm 0.001$
        & $78.486 \pm 0.158$
        & $79.137 \pm 0.142$
        & $0.00 \pm 0.00$ \\
        &
        & Batch shuffle
        & $99.981 \pm 0.015$
        & $78.066 \pm 0.415$
        & $78.720 \pm 0.627$
        & $0.00 \pm 0.00$ \\
        \cmidrule(lr){2-7}

        & \multirow{2}{*}{$(1.50,1.00)$}
        & Aligned
        & $99.966 \pm 0.011$
        & $78.371 \pm 0.208$
        & $79.160 \pm 0.257$
        & $0.00 \pm 0.00$ \\
        &
        & Batch shuffle
        & $99.976 \pm 0.029$
        & $78.278 \pm 0.820$
        & $78.807 \pm 0.751$
        & $0.00 \pm 0.00$ \\
        \cmidrule(lr){2-7}

        & \multirow{2}{*}{$(1.75,1.50)$}
        & Aligned
        & $99.981 \pm 0.008$
        & $78.637 \pm 0.363$
        & $79.477 \pm 0.204$
        & $0.00 \pm 0.00$ \\
        &
        & Batch shuffle
        & $99.965 \pm 0.020$
        & $78.281 \pm 0.372$
        & $78.950 \pm 0.265$
        & $0.00 \pm 0.00$ \\

        \midrule

        \multirow{6}{*}{Mixed}
        & \multirow{2}{*}{$(1.25,1.50)$}
        & Aligned
        & $99.405 \pm 0.007$
        & $79.771 \pm 0.180$
        & $80.107 \pm 0.159$
        & $18.00 \pm 27.78$ \\
        &
        & Batch shuffle
        & $99.391 \pm 0.008$
        & $79.622 \pm 0.403$
        & $80.000 \pm 0.350$
        & $22.00 \pm 38.11$ \\
        \cmidrule(lr){2-7}

        & \multirow{2}{*}{$(1.50,1.00)$}
        & Aligned
        & $98.706 \pm 0.050$
        & $79.832 \pm 0.047$
        & $80.407 \pm 0.133$
        & $14.67 \pm 10.07$ \\
        &
        & Batch shuffle
        & $98.726 \pm 0.062$
        & $79.529 \pm 0.254$
        & $80.070 \pm 0.304$
        & $1.33 \pm 1.15$ \\
        \cmidrule(lr){2-7}

        & \multirow{2}{*}{$(1.75,1.50)$}
        & Aligned
        & $96.908 \pm 0.096$
        & $80.118 \pm 0.227$
        & $80.513 \pm 0.165$
        & $69.33 \pm 44.47$ \\
        &
        & Batch shuffle
        & $97.230 \pm 0.148$
        & $79.281 \pm 0.276$
        & $80.003 \pm 0.201$
        & $0.67 \pm 1.15$ \\

        \bottomrule
    \end{tabular}
\end{table*}

\paragraph{Epoch shuffle.}
We evaluate the three representative configurations for both pure and mixed
replay using Seeds $1$, $42$, and $15213$. For mixed replay,
$(\lambda_{\mathrm{mix}},\sigma_{\mathrm{iso}})=(0.10,0.40)$.
The source SAM epochs are randomly permuted. For SGD training epoch $e$,
the batch-window statistics are taken from the corresponding permuted source
SAM epoch, while the batch-window index within the epoch is unchanged.
Thus, SGD window $j$ continues to use source window $j$, but from a different
SAM epoch. Epoch shuffling therefore changes the epoch-level alignment
between the SAM source trajectory and the SGD training trajectory while
preserving the within-epoch batch-window ordering. The results are presented in Table~\ref{tab:epoch-shuffle-multiseed}.

For pure replay, the effect of epoch shuffling is small and varies across
the three configurations. The mean late-stage validation accuracy changes
from $78.486\%$ to $77.897\%$, from $78.371\%$ to $78.674\%$, and from
$78.637\%$ to $78.626\%$, respectively. Late-stage training accuracy
changes little, and none of the pure replay configurations reaches $80\%$
validation accuracy in the final 50 epochs either before or after epoch
shuffling.

For mixed replay, epoch shuffling reduces the mean late-stage validation
accuracy for all three representative configurations. The changes are from
$79.771\%$ to $79.601\%$, from $79.832\%$ to $79.609\%$, and from
$80.118\%$ to $79.588\%$, respectively. The corresponding Late $N_{80}$
values decrease from $18.00\%$ to $0.67\%$, from $14.67\%$ to $8.00\%$,
and from $69.33\%$ to $0.67\%$. Overall, changing the epoch-level alignment
has a limited and configuration-dependent effect for pure replay, while it
consistently reduces validation performance and high-validation persistence
for the three mixed replay configurations.

\begin{table*}[t]
    \caption{
    Comparison of aligned and epoch-shuffled replay over Seeds $1$, $42$,
    and $15213$. Results are reported as mean $\pm$ sample standard deviation.
    Train Acc.\ and Val.\ Acc.\ are averaged over epochs 201--250.
    Best Val.\ is the maximum validation accuracy over all 250 epochs.
    Late $N_{80}$ is the percentage of epochs among epochs 201--250 with
    validation accuracy $\geq80\%$. For mixed replay,
    $(\lambda_{\mathrm{mix}},\sigma_{\mathrm{iso}})=(0.10,0.40)$.
    }
    \label{tab:epoch-shuffle-multiseed}
    \centering
    \small
    \setlength{\tabcolsep}{3.5pt}
    \begin{tabular}{lllcccc}
        \toprule
        Replay
        & $(\alpha,\beta)$
        & Setting
        & Train Acc.
        & Val. Acc.
        & Best Val.
        & Late $N_{80}$ (\%) \\
        \midrule

        \multirow{6}{*}{Pure}
        & \multirow{2}{*}{$(1.25,1.50)$}
        & Aligned
        & $99.964 \pm 0.001$
        & $78.486 \pm 0.158$
        & $79.137 \pm 0.142$
        & $0.00 \pm 0.00$ \\
        &
        & Epoch shuffle
        & $99.972 \pm 0.022$
        & $77.897 \pm 0.492$
        & $78.540 \pm 0.567$
        & $0.00 \pm 0.00$ \\
        \cmidrule(lr){2-7}

        & \multirow{2}{*}{$(1.50,1.00)$}
        & Aligned
        & $99.966 \pm 0.011$
        & $78.371 \pm 0.208$
        & $79.160 \pm 0.257$
        & $0.00 \pm 0.00$ \\
        &
        & Epoch shuffle
        & $99.985 \pm 0.012$
        & $78.674 \pm 0.566$
        & $79.193 \pm 0.488$
        & $0.00 \pm 0.00$ \\
        \cmidrule(lr){2-7}

        & \multirow{2}{*}{$(1.75,1.50)$}
        & Aligned
        & $99.981 \pm 0.008$
        & $78.637 \pm 0.363$
        & $79.477 \pm 0.204$
        & $0.00 \pm 0.00$ \\
        &
        & Epoch shuffle
        & $99.969 \pm 0.009$
        & $78.626 \pm 0.123$
        & $79.130 \pm 0.141$
        & $0.00 \pm 0.00$ \\

        \midrule

        \multirow{6}{*}{Mixed}
        & \multirow{2}{*}{$(1.25,1.50)$}
        & Aligned
        & $99.405 \pm 0.007$
        & $79.771 \pm 0.180$
        & $80.107 \pm 0.159$
        & $18.00 \pm 27.78$ \\
        &
        & Epoch shuffle
        & $99.388 \pm 0.011$
        & $79.601 \pm 0.056$
        & $80.047 \pm 0.164$
        & $0.67 \pm 1.15$ \\
        \cmidrule(lr){2-7}

        & \multirow{2}{*}{$(1.50,1.00)$}
        & Aligned
        & $98.706 \pm 0.050$
        & $79.832 \pm 0.047$
        & $80.407 \pm 0.133$
        & $14.67 \pm 10.07$ \\
        &
        & Epoch shuffle
        & $98.700 \pm 0.029$
        & $79.609 \pm 0.239$
        & $80.067 \pm 0.296$
        & $8.00 \pm 12.17$ \\
        \cmidrule(lr){2-7}

        & \multirow{2}{*}{$(1.75,1.50)$}
        & Aligned
        & $96.908 \pm 0.096$
        & $80.118 \pm 0.227$
        & $80.513 \pm 0.165$
        & $69.33 \pm 44.47$ \\
        &
        & Epoch shuffle
        & $97.233 \pm 0.123$
        & $79.588 \pm 0.086$
        & $80.130 \pm 0.200$
        & $0.67 \pm 1.15$ \\

        \bottomrule
    \end{tabular}
\end{table*}

\subsection{Epoch-Wise Aggregation}
\label{app:epochwise_method}

For epoch-wise aggregation, we compute the replay statistics over an entire
source SAM epoch rather than over consecutive windows of $K=10$ minibatches.
Let $\Delta_{e,j}$ denote the recorded SAM perturbation for minibatch $j$ in
epoch $e$. We compute the elementwise epoch-level statistics
\begin{equation}
\mu_e
=
\frac{1}{M_e}
\sum_{j=1}^{M_e}
\Delta_{e,j},
\qquad
\sigma_e
=
\sqrt{
\frac{1}{M_e}
\sum_{j=1}^{M_e}
(\Delta_{e,j}-\mu_e)^2
},
\end{equation}
where $M_e$ is the number of minibatches in epoch $e$.

During SGD training epoch $e$, the same pair $(\mu_e,\sigma_e)$ is used for
every minibatch. The random variables in the replay perturbation are still
resampled independently for each minibatch. Epoch-wise aggregation therefore
preserves the epoch-level correspondence between the source SAM trajectory
and the SGD training trajectory, but removes the within-epoch batch-window
information.

We evaluate the three representative configurations for both pure and mixed
replay using Seed $42$. For mixed replay,
$(\lambda_{\mathrm{mix}},\sigma_{\mathrm{iso}})=(0.10,0.40)$.
The results are reported in Table~\ref{tab:epochwise-seed42}.

For pure replay, all three epoch-wise configurations maintain near-perfect
late-stage training accuracy, while late-stage validation accuracy ranges
from $78.202\%$ to $78.645\%$. None reaches $80\%$ validation accuracy in
the final 50 epochs.

Mixed replay gives higher validation accuracy than pure replay for all three
representative configurations. The late-stage validation accuracies are
$79.498\%$, $79.790\%$, and $79.542\%$ for
$(1.25,1.50)$, $(1.50,1.00)$, and $(1.75,1.50)$, respectively.
Only $(1.50,1.00)$ reaches $80\%$ validation accuracy in the final
50 epochs, with Late $N_{80}=6/50$. Compared with the corresponding aligned
batch-window results, epoch-wise aggregation reduces the high-validation
persistence for $(1.25,1.50)$ and $(1.75,1.50)$, while
$(1.50,1.00)$ remains similar. Thus, removing within-epoch batch-window
information generally weakens the validation behavior of mixed replay at
these representative configurations, although the effect depends on the
configuration.

We separately evaluate the mean-only setting
$(\alpha,\beta,\lambda_{\mathrm{mix}},\sigma_{\mathrm{iso}})
=(0,1,0,0)$. It reaches $99.983\%$ late-stage training accuracy,
$78.446\%$ late-stage validation accuracy, and $79.080\%$ Best Val.,
with no epoch reaching $80\%$ validation accuracy.

\begin{table}[t]
    \caption{
    Epoch-wise replay results using Seed 42. Train Acc.\ and Val.\ Acc.\
    are averaged over epochs 201--250. Best Val.\ is the maximum
    validation accuracy over all 250 epochs. Late $N_{80}$ is the number
    of epochs among epochs 201--250 with validation accuracy $\geq80\%$.
    For mixed replay, $(\lambda_{\mathrm{mix}},\sigma_{\mathrm{iso}})
    =(0.10,0.40)$.
    }
    \label{tab:epochwise-seed42}
    \centering
    \begin{tabular}{llcccc}
        \toprule
        Replay
        & $(\alpha,\beta)$
        & Late Train
        & Late Val.
        & Best Val.
        & Late $N_{80}$ \\
        \midrule

        Pure
        & $(1.25,1.50)$
        & 99.995
        & 78.202
        & 78.570
        & 0/50 \\

        Pure
        & $(1.50,1.00)$
        & 99.984
        & 78.645
        & 79.150
        & 0/50 \\

        Pure
        & $(1.75,1.50)$
        & 99.992
        & 78.433
        & 78.970
        & 0/50 \\

        \midrule

        Mixed
        & $(1.25,1.50)$
        & 99.407
        & 79.498
        & 79.940
        & 0/50 \\

        Mixed
        & $(1.50,1.00)$
        & 98.718
        & 79.790
        & 80.180
        & 6/50 \\

        Mixed
        & $(1.75,1.50)$
        & 96.759
        & 79.542
        & 80.150
        & 0/50 \\

        \bottomrule
    \end{tabular}
\end{table}

\subsection{Layer-Wise Intervention}
\label{app:layerwise_method}

To study where the replay perturbation is effective, we restrict perturbation
injection to selected layers while leaving all remaining parameters
unperturbed. We evaluate the three representative configurations for both
pure and mixed replay using Seeds $1$, $42$, and $15213$. For mixed replay,
$(\lambda_{\mathrm{mix}},\sigma_{\mathrm{iso}})=(0.10,0.40)$.
Table~\ref{tab:layerwise-multiseed} reports results for perturbing all
trainable parameters, L1--L4, L1--L4 + FC, and FC only.

For pure replay, changing the perturbed layers has limited effect on
validation performance. Across all three configurations, late-stage
validation accuracy remains between $78.117\%$ and $78.607\%$, and Late
$N_{80}$ remains $0\%$ for every tested layer group. Late-stage training
accuracy remains close to $100\%$ in all cases.

The layer choice has a substantially larger effect for mixed replay.
Perturbing L1--L4 gives higher late-stage validation accuracy than perturbing
all parameters for all three representative configurations. The corresponding
validation accuracies increase from $79.771\%$ to $80.308\%$, from
$79.832\%$ to $80.691\%$, and from $80.118\%$ to $80.177\%$.
Late $N_{80}$ also increases from $18.00\%$ to $80.00\%$ and from
$14.67\%$ to $99.33\%$ for the first two configurations, while remaining
$69.33\%$ for $(1.75,1.50)$.

Adding FC to L1--L4 reduces validation performance relative to L1--L4 for
all three mixed replay configurations. FC-only replay performs substantially
worse, with late-stage validation accuracies of $75.944\%$, $76.412\%$,
and $76.002\%$, while Late $N_{80}$ is $0\%$ in all three cases.
These results show that the effect of layer selection is specific to mixed
replay in the tested configurations, with perturbations applied to L1--L4
giving the strongest validation performance among the reported layer groups.

\begin{table*}[]
    \caption{
    Layer-wise replay results over Seeds $1$, $42$, and $15213$.
    Results are reported as mean $\pm$ sample standard deviation.
    Train Acc.\ and Val.\ Acc.\ are averaged over epochs 201--250.
    Best Val.\ is the maximum validation accuracy over all 250 epochs.
    Late $N_{80}$ is the percentage of epochs among epochs 201--250 with
    validation accuracy $\geq80\%$. For mixed replay,
    $(\lambda_{\mathrm{mix}},\sigma_{\mathrm{iso}})=(0.10,0.40)$.
    }
    \label{tab:layerwise-multiseed}
    \centering
    \footnotesize
    \setlength{\tabcolsep}{3pt}
    \renewcommand{\arraystretch}{0.95}
    \begin{tabular}{lllcccc}
        \toprule
        Replay
        & $(\alpha,\beta)$
        & Layers
        & Late Train
        & Late Val.
        & Best Val.
        & Late $N_{80}$ (\%) \\
        \midrule

        \multirow{12}{*}{Pure}
        & \multirow{4}{*}{$(1.25,1.50)$}
        & All
        & $99.964 \pm 0.001$
        & $78.486 \pm 0.158$
        & $79.137 \pm 0.142$
        & $0.00 \pm 0.00$ \\
        &
        & L1--L4
        & $99.966 \pm 0.013$
        & $78.459 \pm 0.256$
        & $79.230 \pm 0.298$
        & $0.00 \pm 0.00$ \\
        &
        & L1--L4 + FC
        & $99.984 \pm 0.007$
        & $78.497 \pm 0.124$
        & $79.157 \pm 0.064$
        & $0.00 \pm 0.00$ \\
        &
        & FC only
        & $99.982 \pm 0.011$
        & $78.471 \pm 0.148$
        & $79.003 \pm 0.208$
        & $0.00 \pm 0.00$ \\
        \cmidrule(lr){2-7}

        & \multirow{4}{*}{$(1.50,1.00)$}
        & All
        & $99.966 \pm 0.011$
        & $78.371 \pm 0.208$
        & $79.160 \pm 0.257$
        & $0.00 \pm 0.00$ \\
        &
        & L1--L4
        & $99.964 \pm 0.011$
        & $78.607 \pm 0.282$
        & $79.417 \pm 0.306$
        & $0.00 \pm 0.00$ \\
        &
        & L1--L4 + FC
        & $99.978 \pm 0.008$
        & $78.221 \pm 0.111$
        & $78.907 \pm 0.099$
        & $0.00 \pm 0.00$ \\
        &
        & FC only
        & $99.972 \pm 0.0005$
        & $78.245 \pm 0.261$
        & $78.873 \pm 0.301$
        & $0.00 \pm 0.00$ \\
        \cmidrule(lr){2-7}

        & \multirow{4}{*}{$(1.75,1.50)$}
        & All
        & $99.981 \pm 0.008$
        & $78.637 \pm 0.363$
        & $79.477 \pm 0.204$
        & $0.00 \pm 0.00$ \\
        &
        & L1--L4
        & $99.961 \pm 0.004$
        & $78.415 \pm 0.471$
        & $79.133 \pm 0.398$
        & $0.00 \pm 0.00$ \\
        &
        & L1--L4 + FC
        & $99.978 \pm 0.004$
        & $78.246 \pm 0.336$
        & $79.097 \pm 0.292$
        & $0.00 \pm 0.00$ \\
        &
        & FC only
        & $99.973 \pm 0.011$
        & $78.117 \pm 0.502$
        & $78.727 \pm 0.433$
        & $0.00 \pm 0.00$ \\

        \midrule

        \multirow{12}{*}{Mixed}
        & \multirow{4}{*}{$(1.25,1.50)$}
        & All
        & $99.405 \pm 0.007$
        & $79.771 \pm 0.180$
        & $80.107 \pm 0.159$
        & $18.00 \pm 27.78$ \\
        &
        & L1--L4
        & $99.506 \pm 0.010$
        & $80.308 \pm 0.384$
        & $80.817 \pm 0.469$
        & $80.00 \pm 21.07$ \\
        &
        & L1--L4 + FC
        & $99.468 \pm 0.033$
        & $79.726 \pm 0.269$
        & $80.197 \pm 0.242$
        & $19.33 \pm 17.01$ \\
        &
        & FC only
        & $99.968 \pm 0.005$
        & $75.944 \pm 0.116$
        & $76.740 \pm 0.098$
        & $0.00 \pm 0.00$ \\
        \cmidrule(lr){2-7}

        & \multirow{4}{*}{$(1.50,1.00)$}
        & All
        & $98.706 \pm 0.050$
        & $79.832 \pm 0.047$
        & $80.407 \pm 0.133$
        & $14.67 \pm 10.07$ \\
        &
        & L1--L4
        & $99.031 \pm 0.059$
        & $80.691 \pm 0.260$
        & $81.233 \pm 0.217$
        & $99.33 \pm 1.15$ \\
        &
        & L1--L4 + FC
        & $98.937 \pm 0.028$
        & $79.846 \pm 0.232$
        & $80.327 \pm 0.211$
        & $32.67 \pm 32.02$ \\
        &
        & FC only
        & $99.972 \pm 0.006$
        & $76.412 \pm 0.299$
        & $77.167 \pm 0.336$
        & $0.00 \pm 0.00$ \\
        \cmidrule(lr){2-7}

        & \multirow{4}{*}{$(1.75,1.50)$}
        & All
        & $96.908 \pm 0.096$
        & $80.118 \pm 0.227$
        & $80.513 \pm 0.165$
        & $69.33 \pm 44.47$ \\
        &
        & L1--L4
        & $97.983 \pm 0.136$
        & $80.177 \pm 0.369$
        & $80.713 \pm 0.351$
        & $69.33 \pm 48.01$ \\
        &
        & L1--L4 + FC
        & $97.782 \pm 0.071$
        & $79.734 \pm 0.318$
        & $80.197 \pm 0.386$
        & $22.00 \pm 26.91$ \\
        &
        & FC only
        & $99.957 \pm 0.012$
        & $76.002 \pm 0.592$
        & $76.740 \pm 0.443$
        & $0.00 \pm 0.00$ \\

        \bottomrule
    \end{tabular}
\end{table*}

\begin{figure*}[]
    \centering
    \includegraphics[width=\textwidth]{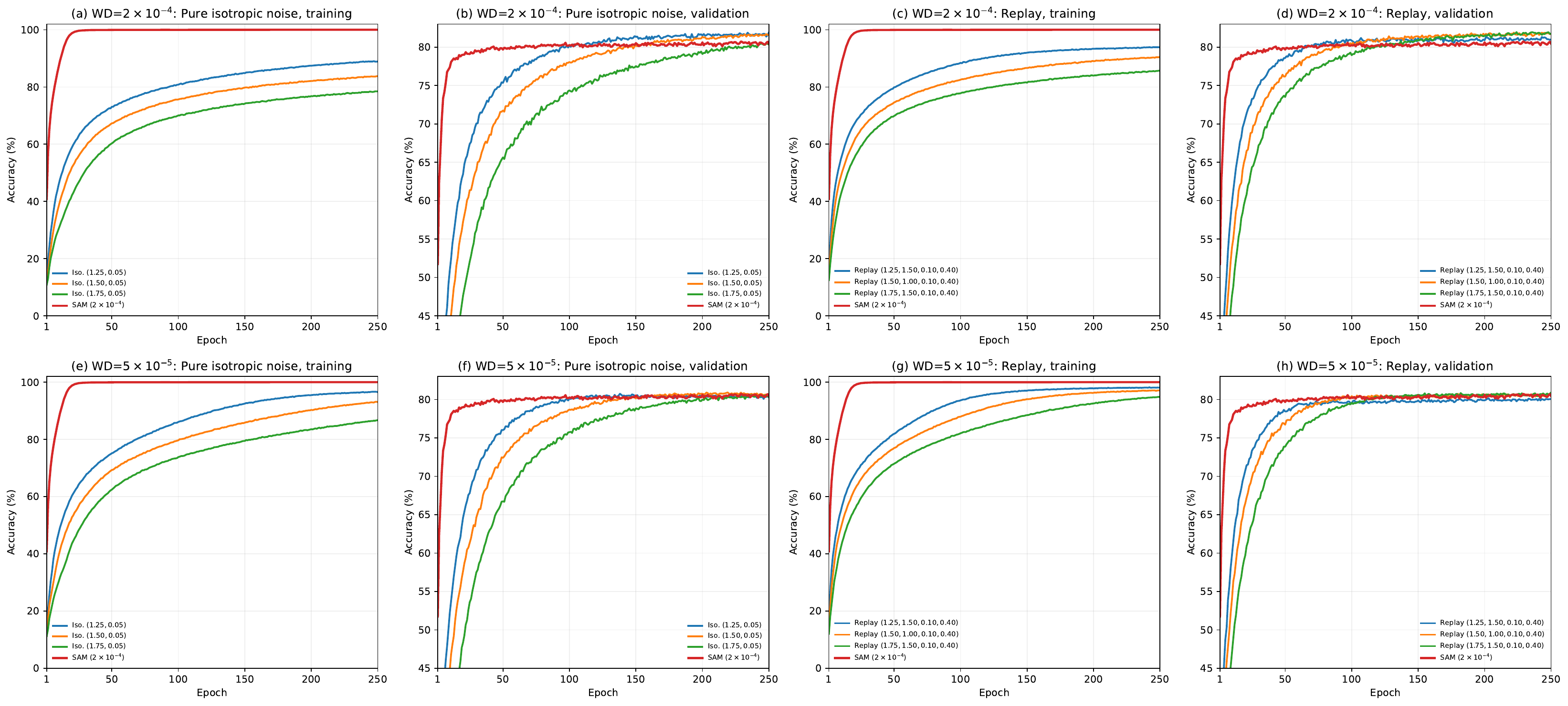}
    \caption{
    Training and validation accuracy trajectories for pure isotropic
    perturbations, statistical replay, and standard SAM under weight decay.
    Replay and isotropic results are shown for weight decay
    $5\times10^{-5}$ and $2\times10^{-4}$. Results are averaged over
    Seeds $1$, $42$, and $15213$.
    }
    \label{fig:weight_decay_trajectories}
\end{figure*}

\begin{table*}[]
    \caption{
    Effect of weight decay on standard SGD, pure isotropic perturbations,
    SAM-derived replay, and standard SAM. Train Acc.\ and Val.\ Acc.\ are
    averaged over epochs 201--250. Best Val.\ is the maximum validation
    accuracy over all 250 epochs. Late $N_{80}$ is the percentage of epochs
    among epochs 201--250 with validation accuracy $\geq80\%$.
    Results are reported as mean $\pm$ standard deviation over Seeds
    $1$, $42$, and $15213$.
    }
    \label{tab:weight-decay-comparison}
    \centering
    \footnotesize
    \setlength{\tabcolsep}{4pt}
    \renewcommand{\arraystretch}{0.99}
    \begin{tabular}{lccccc}
        \toprule
        Method
        & WD
        & Late Train
        & Late Val.
        & Best Val.
        & Late $N_{80}$ (\%) \\
        \midrule

        Standard SAM
        & $2\times10^{-4}$
        & $99.996 \pm 0.004$
        & $80.474 \pm 0.208$
        & $81.000 \pm 0.240$
        & $93.33 \pm 5.77$ \\

        \midrule

        \multirow{3}{*}{SGD}
        & $0$
        & $99.959 \pm 0.011$
        & $78.077 \pm 0.291$
        & $78.927 \pm 0.285$
        & $0.00 \pm 0.00$ \\
        & $5\times10^{-5}$
        & $99.795 \pm 0.186$
        & $78.421 \pm 0.758$
        & $79.460 \pm 0.217$
        & $0.00 \pm 0.00$ \\
        & $2\times10^{-4}$
        & $99.032 \pm 0.010$
        & $76.693 \pm 0.252$
        & $77.823 \pm 0.300$
        & $0.00 \pm 0.00$ \\

        \midrule

        \multirow{3}{*}{Pure Iso $(1.25,0.05)$}
        & $0$
        & $98.483 \pm 0.035$
        & $79.523 \pm 0.296$
        & $80.043 \pm 0.271$
        & $6.67 \pm 11.55$ \\
        & $5\times10^{-5}$
        & $96.169 \pm 0.026$
        & $80.374 \pm 0.248$
        & $80.937 \pm 0.287$
        & $93.33 \pm 11.55$ \\
        & $2\times10^{-4}$
        & $88.319 \pm 0.284$
        & $81.616 \pm 0.165$
        & $82.037 \pm 0.130$
        & $100.00 \pm 0.00$ \\
        \cmidrule(lr){2-6}

        \multirow{3}{*}{Pure Iso $(1.50,0.05)$}
        & $0$
        & $95.665 \pm 0.241$
        & $80.109 \pm 0.290$
        & $80.573 \pm 0.240$
        & $64.00 \pm 40.60$ \\
        & $5\times10^{-5}$
        & $91.854 \pm 0.047$
        & $80.728 \pm 0.023$
        & $81.077 \pm 0.025$
        & $100.00 \pm 0.00$ \\
        & $2\times10^{-4}$
        & $82.999 \pm 0.400$
        & $81.395 \pm 0.148$
        & $81.873 \pm 0.136$
        & $100.00 \pm 0.00$ \\
        \cmidrule(lr){2-6}

        \multirow{3}{*}{Pure Iso $(1.75,0.05)$}
        & $0$
        & $89.845 \pm 0.711$
        & $79.610 \pm 0.281$
        & $80.040 \pm 0.294$
        & $12.67 \pm 21.94$ \\
        & $5\times10^{-5}$
        & $85.194 \pm 0.592$
        & $80.284 \pm 0.078$
        & $80.683 \pm 0.100$
        & $90.67 \pm 5.77$ \\
        & $2\times10^{-4}$
        & $77.665 \pm 0.471$
        & $79.916 \pm 0.191$
        & $80.703 \pm 0.310$
        & $45.33 \pm 19.43$ \\

        \midrule

        \multirow{3}{*}{Replay $(1.25,1.50,0.10,0.40)$}
        & $0$
        & $99.405 \pm 0.007$
        & $79.771 \pm 0.180$
        & $80.107 \pm 0.159$
        & $18.00 \pm 27.79$ \\
        & $5\times10^{-5}$
        & $98.021 \pm 0.005$
        & $79.961 \pm 0.251$
        & $80.370 \pm 0.282$
        & $44.67 \pm 41.10$ \\
        & $2\times10^{-4}$
        & $93.631 \pm 0.147$
        & $81.059 \pm 0.036$
        & $81.603 \pm 0.127$
        & $100.00 \pm 0.00$ \\
        \cmidrule(lr){2-6}

        \multirow{3}{*}{Replay $(1.50,1.00,0.10,0.40)$}
        & $0$
        & $98.706 \pm 0.050$
        & $79.832 \pm 0.047$
        & $80.407 \pm 0.133$
        & $14.67 \pm 10.07$ \\
        & $5\times10^{-5}$
        & $96.790 \pm 0.070$
        & $80.560 \pm 0.111$
        & $81.020 \pm 0.113$
        & $100.00 \pm 0.00$ \\
        & $2\times10^{-4}$
        & $89.806 \pm 0.083$
        & $81.660 \pm 0.197$
        & $82.063 \pm 0.183$
        & $100.00 \pm 0.00$ \\
        \cmidrule(lr){2-6}

        \multirow{3}{*}{Replay $(1.75,1.50,0.10,0.40)$}
        & $0$
        & $96.908 \pm 0.096$
        & $80.118 \pm 0.227$
        & $80.513 \pm 0.165$
        & $69.33 \pm 44.47$ \\
        & $5\times10^{-5}$
        & $93.886 \pm 0.091$
        & $80.647 \pm 0.346$
        & $81.063 \pm 0.359$
        & $96.67 \pm 5.77$ \\
        & $2\times10^{-4}$
        & $84.927 \pm 0.363$
        & $81.695 \pm 0.376$
        & $82.067 \pm 0.375$
        & $100.00 \pm 0.00$ \\

        \bottomrule
    \end{tabular}
\end{table*}

\subsection{Effect of weight decay.}
\label{app:weight_decay_results}
We evaluate weight decay values of $0$, $5\times10^{-5}$, and
$2\times10^{-4}$ for SGD, statistical replay, and pure isotropic
perturbations in Table~\ref{tab:weight-decay-comparison}. Results are
averaged over Seeds $1$, $42$, and $15213$. For most replay and isotropic
configurations, increasing weight decay reduces late-stage training accuracy
while improving late-stage validation accuracy and Late $N_{80}$. At weight
decay $2\times10^{-4}$, five of the six perturbation configurations reach
Late $N_{80}=100\%$, with late-stage validation accuracies between
$81.059\%$ and $81.695\%$. For these five configurations, late-stage
training accuracies range from $82.999\%$ to $93.631\%$, compared with
$99.996\%$ for standard SAM.

The largest isotropic perturbation setting, $(1.75,0.05)$, is an exception.
Increasing weight decay from $5\times10^{-5}$ to $2\times10^{-4}$ reduces
its late-stage validation accuracy from $80.284\%$ to $79.916\%$ and Late
$N_{80}$ from $90.67\%$ to $45.33\%$. Figure~\ref{fig:weight_decay_trajectories}
also shows that replay and isotropic training accuracy increases more slowly
than standard SAM as weight decay increases, while their late-stage
validation performance remains comparable. In contrast, standard SGD has
Late $N_{80}=0\%$ for all three tested weight decay values. These results
show that weight decay changes the tradeoff between training fit and
validation performance for replay and isotropic perturbations, but does not
reproduce SAM's combination of near-perfect training fit and high validation
performance.


\section{Curvature-Regularization Baselines}
\label{app:curvature}


We use preliminary curvature baselines to test whether direct
regularization of classifier head curvature can reproduce SAM in the
setting studied here. We consider a trace based RF style regularizer~\citep{petzka2021relative,adilova2023fam,han2025flatness,walter2026when} and
a spectral variant based on the largest eigenvalue of the classifier
head Hessian. Both quantities are computed from the current minibatch
and differentiated through the full network.

\subsection{Trace-Based RF-Style Regularization}

Let $W$ denote the classifier-head weight matrix and let
\[
H_W = \nabla_W^2 \mathcal{L}_{\mathrm{CE}}
\]
denote the Hessian of the minibatch cross-entropy loss with respect to $W$.
We use
\begin{equation}
R_{\mathrm{tr}}
=
\|W\|_F^2 \operatorname{Tr}(H_W).
\label{eq:trace-rf}
\end{equation}

For a minibatch of size $B$, with penultimate representation $\phi_b$ and
softmax probabilities $p_{b,c}$, the trace used in our implementation is
\begin{equation}
\operatorname{Tr}(H_W)
=
\frac{1}{B}
\sum_{b=1}^{B}
\left[
    \sum_c p_{b,c}(1-p_{b,c})
\right]
\|\phi_b\|_2^2.
\label{eq:trace-head}
\end{equation}
We do not normalize $\phi_b$, detach $\|W\|_F$, or apply a logarithmic
transformation to $R_{\mathrm{tr}}$.

\paragraph{Fixed coefficient.}
The first variant uses
\begin{equation}
\mathcal{L}_{\mathrm{fixed}}
=
\mathcal{L}_{\mathrm{CE}}
+
10^{-4} R_{\mathrm{tr}}.
\label{eq:trace-fixed}
\end{equation}
The regularizer is active from the first epoch and is optimized by standard
backpropagation.

\paragraph{Adaptive coefficient with gradient clipping.}
We also test an adaptive variant. For each minibatch, we set
\begin{equation}
\lambda_t
=
\frac{
    \operatorname{stopgrad}(\mathcal{L}_{\mathrm{CE}})
}{
    \operatorname{stopgrad}(R_{\mathrm{tr}}) + 10^{-8}
}.
\label{eq:trace-dynamic-lambda}
\end{equation}
This makes the scalar value of $\lambda_t R_{\mathrm{tr}}$ approximately
equal to $\mathcal{L}_{\mathrm{CE}}$. The coefficient $\lambda_t$ is treated
as constant when computing gradients.

Let
\[
g_{\mathrm{CE}}
=
\nabla \mathcal{L}_{\mathrm{CE}},
\qquad
g_{\mathrm{tr}}
=
\lambda_t \nabla R_{\mathrm{tr}}.
\]
We limit the norm of the trace-regularization gradient to at most $30\%$ of
the cross-entropy gradient norm:
\begin{equation}
s_t
=
\min\left(
1,
\frac{
    0.3\|g_{\mathrm{CE}}\|_2
}{
    \|g_{\mathrm{tr}}\|_2 + 10^{-12}
}
\right).
\label{eq:trace-clip}
\end{equation}
The optimizer update uses
\begin{equation}
g_t
=
g_{\mathrm{CE}} + s_t g_{\mathrm{tr}}.
\label{eq:trace-update}
\end{equation}

\subsection{Spectral RF-Style Regularization}

We next replace the Hessian trace by its largest eigenvalue. The spectral
regularizer is
\begin{equation}
R_{\mathrm{spec}}
=
\|W\|_F^2 \lambda_{\max}(H_W),
\label{eq:spectral-rf}
\end{equation}
and the training objective is
\begin{equation}
\mathcal{L}_{\mathrm{spec}}
=
\mathcal{L}_{\mathrm{CE}}
+
\lambda_{\mathrm{spec}} R_{\mathrm{spec}}.
\label{eq:spectral-objective}
\end{equation}
Our main spectral experiment uses
\[
\lambda_{\mathrm{spec}} = 10^{-4}.
\]
The largest eigenvalue is recomputed from the current logits and penultimate
representations for every minibatch. No adaptive weighting or gradient
clipping is used for this baseline.

\subsection{Results}

Table~\ref{tab:curvature-baselines} summarizes the classifier head
curvature baselines. We treat the reported coefficients and gradient cap as representative preliminary settings rather than the result of an exhaustive hyperparameter search. For reference, standard SGD and SAM obtain mean
late validation accuracies of $78.077\%$ and $80.474\%$, respectively,
as reported in Table~\ref{tab:main-results}. All three curvature
controls obtain higher mean late validation accuracy than SGD but
remain closer to SGD than to SAM. Spectral regularization obtains the
highest value of $78.595\pm0.181\%$, which remains $1.879$ percentage
points below SAM.

Across the complete trajectories, validation accuracy reaches at least
$79\%$ in 84, 21, and 66 of the 750 seed epochs for fixed trace,
adaptive trace, and spectral regularization, respectively, compared
with 3 epochs for SGD. These occurrences are seed dependent: fixed
trace reaches this threshold only for Seed 15213, adaptive trace only
for Seed 1, and spectral regularization for Seeds 1 and 42. Moreover,
Late $N_{80}$ is zero for all three controls. Thus, the tested
classifier head curvature regularizers show modest but seed dependent
improvements over SGD and are insufficient to reproduce SAM in this
setting.

\begin{table}[t]
    \caption{
    Classifier-head curvature baselines on CIFAR-10 with ResNet-18 without
    data augmentation. Late Train and Late Val.\ are averaged over epochs
    201--250. Best Val.\ is the maximum validation accuracy over all
    250 epochs. Late $N_{80}$ is the percentage of epochs among epochs
    201--250 with validation accuracy $\geq80\%$.
    Results are reported as mean $\pm$ sample standard deviation over
    Seeds $1$, $42$, and $15213$.
    }
    \label{tab:curvature-baselines}
    \centering
    \small
    \setlength{\tabcolsep}{4pt}
    \begin{tabular}{lcccc}
        \toprule
        Method
        & Late Train
        & Late Val.
        & Best Val.
        & Late $N_{80}$ (\%) \\
        \midrule

        Fixed Trace ($\lambda_{\mathrm{tr}}=10^{-4}$)
        & $99.974 \pm 0.019$
        & $78.511 \pm 0.456$
        & $79.213 \pm 0.577$
        & $0.00 \pm 0.00$ \\

        Adaptive Trace (30\% cap)
        & $99.756 \pm 0.025$
        & $78.438 \pm 0.383$
        & $78.987 \pm 0.325$
        & $0.00 \pm 0.00$ \\

        Spectral ($\lambda_{\mathrm{spec}}=10^{-4}$)
        & $99.975 \pm 0.005$
        & $78.595 \pm 0.181$
        & $79.107 \pm 0.251$
        & $0.00 \pm 0.00$ \\

        \bottomrule
    \end{tabular}
\end{table}




\newpage
\section*{NeurIPS Paper Checklist}


\begin{enumerate}

\item {\bf Claims}
    \item[] Question: Do the main claims made in the abstract and introduction accurately reflect the paper's contributions and scope?
    \item[] Answer: \answerYes{} 
    \item[] Justification: We provide detailed results for the claims in both introduction and abstract.

\item {\bf Limitations}
    \item[] Question: Does the paper discuss the limitations of the work performed by the authors?
    \item[] Answer: \answerYes{} 
    \item[] Justification: We have a limitation discussion in Section 4.

\item {\bf Theory assumptions and proofs}
    \item[] Question: For each theoretical result, does the paper provide the full set of assumptions and a complete (and correct) proof?
    \item[] Answer: \answerNA{} 
    \item[] Justification: Not related to theory.

    \item {\bf Experimental result reproducibility}
    \item[] Question: Does the paper fully disclose all the information needed to reproduce the main experimental results of the paper to the extent that it affects the main claims and/or conclusions of the paper (regardless of whether the code and data are provided or not)?
    \item[] Answer: \answerYes{} 
    \item[] Justification: We include all the implementation details in appendix.

\item {\bf Open access to data and code}
    \item[] Question: Does the paper provide open access to the data and code, with sufficient instructions to faithfully reproduce the main experimental results, as described in supplemental material?
    \item[] Answer: \answerNo{} 
    \item[] Justification: We use the publicly available CIFAR-10 dataset and provide implementation details in the appendix. The code will be released after publication. 

\item {\bf Experimental setting/details}
    \item[] Question: Does the paper specify all the training and test details (e.g., data splits, hyperparameters, how they were chosen, type of optimizer) necessary to understand the results?
    \item[] Answer: \answerYes{} 
    \item[] Justification: We have all required implementation details in appendix.

\item {\bf Experiment statistical significance}
    \item[] Question: Does the paper report error bars suitably and correctly defined or other appropriate information about the statistical significance of the experiments?
    \item[] Answer: \answerYes{} 
    \item[] Justification: Yes, we run experiments for three random seeds for all necessary experiments.

\item {\bf Experiments compute resources}
    \item[] Question: For each experiment, does the paper provide sufficient information on the computer resources (type of compute workers, memory, time of execution) needed to reproduce the experiments?
    \item[] Answer: \answerYes{} 
    \item[] Justification: We specify the type of GPU we use to conduct experiments.
    
\item {\bf Code of ethics}
    \item[] Question: Does the research conducted in the paper conform, in every respect, with the NeurIPS Code of Ethics \url{https://neurips.cc/public/EthicsGuidelines}?
    \item[] Answer: \answerYes{} 
    \item[] Justification: We have read the code and we honor the code.

\item {\bf Broader impacts}
    \item[] Question: Does the paper discuss both potential positive societal impacts and negative societal impacts of the work performed?
    \item[] Answer: \answerNA{} 
    \item[] Justification: We do not use sensitive data.
    
\item {\bf Safeguards}
    \item[] Question: Does the paper describe safeguards that have been put in place for responsible release of data or models that have a high risk for misuse (e.g., pre-trained language models, image generators, or scraped datasets)?
    \item[] Answer: \answerNA{} 
    \item[] Justification: Not related to our paper topic.

\item {\bf Licenses for existing assets}
    \item[] Question: Are the creators or original owners of assets (e.g., code, data, models), used in the paper, properly credited and are the license and terms of use explicitly mentioned and properly respected?
    \item[] Answer: \answerNA{} 
    \item[] Justification: We use public dataset and model.

\item {\bf New assets}
    \item[] Question: Are new assets introduced in the paper well documented and is the documentation provided alongside the assets?
    \item[] Answer: \answerNA{} 
    \item[] Justification: We use public dataset and model.

\item {\bf Crowdsourcing and research with human subjects}
    \item[] Question: For crowdsourcing experiments and research with human subjects, does the paper include the full text of instructions given to participants and screenshots, if applicable, as well as details about compensation (if any)? 
    \item[] Answer: \answerNA{} 
    \item[] Justification: No crowdsourcing or human-subject research is involved.

\item {\bf Institutional review board (IRB) approvals or equivalent for research with human subjects}
    \item[] Question: Does the paper describe potential risks incurred by study participants, whether such risks were disclosed to the subjects, and whether Institutional Review Board (IRB) approvals (or an equivalent approval/review based on the requirements of your country or institution) were obtained?
    \item[] Answer: \answerNA{} 
    \item[] Justification: No crowdsourcing or human-subject research is involved.

\item {\bf Declaration of LLM usage}
    \item[] Question: Does the paper describe the usage of LLMs if it is an important, original, or non-standard component of the core methods in this research? Note that if the LLM is used only for writing, editing, or formatting purposes and does \emph{not} impact the core methodology, scientific rigor, or originality of the research, declaration is not required.
    \item[] Answer: \answerNA{} 
    \item[] Justification: LLMs were used only for language editing and did not affect the methodology, experiments, or scientific conclusions.

\end{enumerate}

\end{document}